\documentclass{article}
\usepackage{spconf}

\usepackage{amsmath,amssymb,amsthm}
\usepackage{graphicx}
\usepackage{subfigure}
\usepackage{caption}
\usepackage{float}
\usepackage{bm}

\usepackage{algorithm}
\usepackage{algorithmicx}
\usepackage{algpseudocode}

\usepackage{indentfirst}
\usepackage{color}

\newcommand{\x}		{\mathbf{x}}	
\newcommand{\y}		{\mathbf{y}}
\newcommand{\A}		{\mathbf{A}}
\newcommand{\z}	 {\mathbf{z}}
\newcommand{\W}	 {\mathbf{W}}
\renewcommand{\P}	 {\mathbf{P}}
\newcommand{\omg}	 {\mathbf{\Omega}}
\newcommand{\Y}		{\mathbf{Y}}
\newcommand{\Z}		{\mathbf{Z}}

\newcommand{\D}	 {\mathbf{D}}
\newcommand{\I}	     {\mathbf{I}}

\newcommand{\s}	 {\mathbf{s}}
\newcommand{\g}	 {\mathbf{g}}
\newcommand{\h}	 {\mathbf{h}}
\newcommand{\ze}	 {\bm{\zeta}}

\newcommand{\R}	     {\mathsf{R}}

\newcommand{\diag}  {\mathsf{diag}}

\title{Low Dose CT Image Reconstruction With Learned Sparsifying Transform}

\name{Xuehang Zheng$^1$, Zening Lu$^1$, Saiprasad Ravishankar$^2$, Yong Long*$^{1}$, Jeffrey A. Fessler$^2$ \thanks{This work was supported in part by SJTU-UM Collaborative Research Program, NSFC (61501292), Shanghai Pujiang Talent Program (15PJ1403900), NIH grant U01 EB018753, ONR grant N00014-15-1-2141, DARPA Young Faculty Award D14AP00086, and ARO MURI grants W911NF-11-1-0391 and 2015-05174-05.  \qquad  \qquad \qquad \qquad \qquad \qquad *Yong Long (email:  \texttt{yong.long@sjtu.edu.cn}). }    }

\address{$^1$University of Michigan - Shanghai Jiao Tong University Joint Institute,\\
       Shanghai Jiao Tong University, Shanghai, China\\
       $^2$Department of Electrical Engineering and Computer Science, University of Michigan, MI, USA       
       }

\copyrightnotice{978-1-5090-0746-2/16/\$31.00 {\copyright}2016 IEEE}

\begin{document}
%
\maketitle
\begin{abstract}
A major challenge in computed tomography (CT) is to reduce X-ray dose to a low or even ultra-low level while maintaining the high quality of reconstructed images. We propose a new method for CT reconstruction that combines penalized weighted-least squares reconstruction (PWLS) with regularization based on a sparsifying transform (PWLS-ST) learned from a dataset of numerous CT images. We adopt an alternating algorithm to optimize the PWLS-ST cost function that alternates between a CT image update step and a sparse coding step. We adopt a relaxed linearized augmented Lagrangian method with ordered-subsets (relaxed OS-LALM) to accelerate the CT image update step by reducing the number of forward and backward projections. Numerical experiments on the XCAT phantom show that for low dose levels, the proposed \mbox{PWLS-ST} method dramatically improves the quality of reconstructed images compared to PWLS reconstruction with a nonadaptive edge-preserving regularizer (PWLS-EP).  

\end{abstract}
\section{Introduction}
\label{sec:intro}

A major challenge in computed tomography (CT) is to reduce X-ray dose to a low or even ultra-low level while maintaining the high quality
of reconstructed images. Low dose CT (LDCT) scans that still provide good image quality could significantly improve the benefits of CT scans and open up numerous entirely new clinical applications. 

Currently, most commercial CT scanners use a technique called filtered back-projection (FBP) for image reconstruction. FBP requires undesirably high doses of radiation to produce high-quality diagnostic images. Model-based image reconstruction (MBIR) methods, also known as statistical image reconstruction methods, produce high-quality and accurate images, while reducing patient radiation exposure. Weighted-least squares (WLS) estimation is commonly used for CT \cite{thibault:07:atd}. WLS estimation with proper weighting that gives less weight to measurements that are noisier and more weight to the more reliable data reduces noise in the reconstructed image. Penalized weighted-least squares (PWLS) reconstruction with added regularization based on prior knowledge of the underlying unknown object improves image quality in  LDCT reconstruction \cite{pfister:14:ast}. Thus, MBIR with better image priors is a promising way to develop improved reconstruction methods for achieving high quality LDCT imaging.

Prior information extracted from big datasets of CT images could potentially enable dramatic improvements in image reconstruction from LDCT measurements. It is well known that natural signals are sparse in certain transform domains, such as wavelets and discrete gradient domain. A sparsifying transform (ST)  converts signals into these domains where they can be represented using a few non-zero coefficients. Ravishankar and Bresler \cite{ravishankar:13:lst,ravishankar:15:lst} proposed a generalized analysis dictionary learning method, called transform learning, to efficiently find sparse representations of data. The  transform learning method avoids optimization of highly non-convex or NP-hard cost functions involved in both synthesis \cite{elad:06:idv,aharon:06:ksa} and previous analysis \cite{rubinstein:13:aka} dictionary learning methods, and shows promising performance and speed-ups over the popular synthesis K-SVD \cite{aharon:06:ksa} algorithm in applications such as image denoising. 

Xu et al. \cite{xu:12:ldx} first applied dictionary learning to CT image reconstruction by proposing a PWLS approach with regularization based on a redundant synthesis dictionary. Their method uses a global dictionary trained from image patches extracted from one normal-dose FBP image, or an adaptive dictionary jointly estimated with the low-dose image. Pfister and Bresler \cite{pfister:14:ast,pfister:14:mbi} proposed a model-based iterative reconstruction method with adaptive sparsifying transforms to jointly estimate the ST and the image, showing the promise of PWLS reconstruction with ST regularization. Existing CT image reconstruction methods based on dictionary/transform learning have two common downsides. Firstly, the dictionary is often learned from a very small number of prior images or from the current measurements themselves, which does not take advantage of the existing big databases of CT images acquired from thousands of patients. Secondly, the prior images are often from the same patient at regular dose. When such a prior image is not available, these methods could not be used.

We propose a new method for low dose CT reconstructions that combines conventional PWLS reconstruction with regularization based on a sparsifying transform (PWLS-ST) learned from a dataset of numerous CT images.  Numerical experiments on the XCAT phantom show that for low dose levels, our method dramatically improves the quality of reconstructed images compared to PWLS reconstruction with an edge-preserving hyperbola regularizer (PWLS-EP).  We adopt a relaxed linearized augmented Lagrangian method with ordered-subsets (relaxed OS-LALM) \cite{nien:16:rla} to accelerate the  image reconstruction process.

\section{Problem Formulation}
\label{sec:formulation}

 We solve the following optimization problem to reconstruct an image $\x \in \mathbb{R}^{N_p}$ from noisy sinogram data $\y \in \mathbb{R}^{N_d}$ using a pre-learned \cite{ravishankar:15:lst} ST matrix $\omg \in \mathbb{R}^{ l \times l}$: 
	\begin{equation}\label{eq:P0}	
		 \min_{\x \in \mathcal{C}}  \frac{1}{2}\|\y - \A \x\|^2_{\W}   
		+  \R(\x)   
	\tag{P0}
	\end{equation}
	where the regularizer $\R$ based on the sparsifying transform $\omg$ is defined as:
	\begin{equation}\label{eq:Rx}
	\R(\x) \triangleq   \min_{\{\z_j\}}  \beta \sum_{j=1}^{N}  \bigg\{  \|\omg \P_j \x - \z_{j}\|^2_2 + \gamma^2\|\z_{j}\|_0 \bigg\}  
	\end{equation}			
 $\W = \diag \{w_i\}$ is the diagonal weighting matrix with elements being the reciprocal of the variance, i.e., $w_i \approx 1/\sigma^2(y_i)$, $\A \in  \mathbb{R}^{  N_d  \times N_p}$ is the system matrix of a CT scan, and $\mathcal{C} \triangleq \{\x|x_j \geq 0 , 1 \leq j \leq N_p \} $. $N$ is the number of image patches, the operator $\P_j \in \mathbb{R}^{l\times N_p}$ extracts the $j$th patch of $l$ voxels of $\x$ as a vector $\P_j\x$, $\z_{j} \in \mathbb{R}^{l}$ denotes the sparse representation of $\P_j\x$, $\beta$ is a positive parameter to control the noise and resolution trade-off, $\gamma$ is a weight to control sparsity in the model, and $\|\cdot\|_0$ is the $\ell_0$ ``norm'' that counts the number of nonzero elements in a vector.
	
	In \eqref{eq:P0}, the patches of the underlying image are assumed to be approximately sparse in the learned transform $\omg$ domain. We estimate both the image $\x$ and the sparse coefficients $\{\z_j\}$ from LDCT data.

	\section{Algorithm}
	\label{sec:algorithm}
   
    \subsection{ Sparsifying Transform (ST) Learning}
   
     We learn a sparsifying transform (ST) matrix $\omg$ from the patches extracted from a dataset of regular dose CT images, i.e., we solve the following problem:
    \begin{equation}\label{eq:P1}
    \min_{\omg,\Z}\| \omg\Y-\Z \|_{F}^2 + \lambda  \big( \| \omg \|_{F}^2 - \log|\det \omg| \big) + \sum_{i=1}^{N'} \eta^2\|\Z_{i}\|_0
	\tag{P1}
   	\end{equation}		
	where  $N'$ is the number of training patches, $\Z \in  \mathbb{R}^{l\times N'}$ is a matrix whose columns $\{\Z_i\}$ are the sparse codes of the corresponding training signals (vectorized patches) in $\Y  \in \mathbb{R}^{l\times N'}$, $\lambda$ and $\eta$ are positive scalar parameters, and regularizer  $\| \omg \|_{F}^2 - \log|\det \omg|$ prevents trivial solutions and enables control over the condition number of $\omg$.

    \subsection{Optimization algorithm}
	
	We propose an alternating algorithm to solve \eqref{eq:P0} that alternates between updating $\x$ (\textit{image update step}) and $ \{\z_j\}$ (\textit{sparse coding step}) with other variables kept fixed.  	
	 
	 \subsubsection{ Image Update Step}
	 
	  With $ \{\z_j\}$ fixed, \eqref{eq:P0} reduces to the following weighted least squares problem:
	\begin{equation}
	\min_{\x \in \mathcal{C}} \bigg\{ \frac{1}{2} \|\y - \A\x \|^2_{\W} +  \beta \sum_{j=1}^{N}    \|\omg \P_j \x - \z_{j}\|^2_2   \bigg\} 
	\end{equation}
		   	
	 We solve this problem using the relaxed OS-LALM \cite{nien:16:rla} by iterating over the following update steps:
	 	\begin{equation}\label{eq:rlalm}
	 	\left\{
	 	\begin{aligned}	 		
 		\s^{(k+1)} &= \rho(\D_\A \x^{(k)} -\h^{(k)}) + (1-\rho)\g^{(k)} \\
 		\x^{(k+1)} &= [\x^{(k)} - (\rho\D_\A+\D_\R)^{-1}(\s^{(k+1)} +\nabla \R(\x^{(k)}))]_\mathcal{C} \\
 		\ze^{(k+1)} & \triangleq  M  \A_m' \W_m(\A_m\x^{(k+1)}-\y_m) \\
 		\g^{(k+1)} &= \frac{\rho}{\rho+1}(\alpha \ze^{(k+1)} + (1-\alpha)\g^{(k)}) +  \frac{1}{\rho+1}\g^{(k)}\\
 		\h^{(k+1)}  &= \alpha(\D_{\A} \x^{(k+1)} -\ze^{(k+1)}) + (1-\alpha)\h^{(k)}		
        \end{aligned}	 		
	 	\right.
	 	\end{equation}	
	 	  where $\D_{\A}$ is a diagonal majorizing matrix of $\A'\W\A$, e.g., $\D_{\A} \triangleq \diag \{\A'\W\A\mathbf{1}\} \succeq \A'\W\A$ \cite{erdogan:99:osa}, $[\cdot]_\mathcal{C}$ is an operator that projects the input vector onto the convex set $\mathcal{C}$, $M$ is the number of ordered subsets, and $\A_m$, $\W_m$, and the vector $\y_m$ are sub-matrices of $\A$, $\W$, and sub-vector of $\y$, respectively, for the $m$th subset. $1\leq\alpha<2$ is the \mbox{(over-)relaxation} parameter, and $\rho >0 $ is the AL penalty parameter decreasing gradually as iterations progress \cite{nien:16:rla}, i.e.,
			\begin{equation}\label{eq:rho}		
			\rho_n(\alpha)= \begin{cases}
			1      &,  n=0\\ 
			\frac{\pi}{\alpha(n+1)}\sqrt{1-\big(\frac{\pi}{2\alpha(n+1)}\big)^2}     &,  \text{otherwise.}
			\end{cases}
			\end{equation}		 	 	 
	 	The learned sparsifying transform $\omg$ is a well-conditioned square matrix \cite{ravishankar:15:lst}. $\D_\R \succeq  \nabla^2 \R(\x)= 2\beta \sum_{j=1}^{N}\P_j' \omg' \omg \P_j 	$ is a diagonal majorizing matrix of the Hessian of the regularizer $\R(\x)$, e.g.,
	\begin{equation}
	\D_\R  \triangleq 2 \beta \sum_{j=1}^{N}\P_j'\P_j  \lambda_{\max}(\omg' \omg)  \succeq \nabla^2 \R(\x) 
	\end{equation}

	The term $\sum_{j=1}^{N}\P_j'\P_j \in \mathbb{C}^{N_p \times N_p } $ is a diagonal matrix with the diagonal entries corresponding to image pixel locations and their values being the number of patches overlapping each pixel \cite{ravishankar:11:mir}. If we assume periodically positioned overlapping image patches that wrap around at image boundaries, then the diagonal entries are equal, i.e., $\sum_{j=1}^{N}\P_j'\P_j = \kappa\I $ ($\I \in \mathbb{C}^{N_p \times N_p }$), where $\kappa$ is a scalar. In particular, when the overlap stride is 1, $\kappa$ is equal to the image patch size $l$. Therefore, $\D_\R$ simplifies to:
	\begin{equation}
	\D_\R = 2\beta l \lambda_{\max} (\omg' \omg)  \I
	\end{equation}
	Since $\D_\R$ is independent of $\x$ and $\z_j$, we precompute it prior to iterating.

	 \subsubsection{ Sparse Coding Step}
	 
	  With $\x$ fixed,  we update $\{\z_j\}$ by solving
	 \begin{equation}
	 \min_{\{\z_j\}}  \sum_{j=1}^{N} \bigg\{  \|\omg \P_j \x - \z_{j}\|^2_2 + \gamma^2\|\z_{j}\|_0 \bigg\} 
	 \end{equation} 
     The optimal sparse codes are given in closed form as $\hat{\z}_j =  H_{\gamma}(\omg \P_j \x)$ $\forall \, j$, i.e., setting the entries with magnitude less than $\gamma$ to zero. The \textit{hard-thresholding} operator $\mathit{H}_{\gamma}(\cdot)$ is applied to each entry $b$ in a vector as
	\begin{eqnarray}
	\mathit{H}_{\gamma}(b) \triangleq \begin{cases}
		0,      &|b|   < \gamma \\
		b,   &|b|  \geq \gamma.
	\end{cases}
	\end{eqnarray}

\begin{algorithm}[h]  
	\caption{PWLS-ST Algorithm}\label{alg: PWLS-ST Algorithm}
	\begin{algorithmic}[0]
		\State \textbf{Input:}
		initial image $\tilde{\x}^{(0)}$, $\alpha = 1.999$ , $\rho = 1$, \\
		$\D_{\A} = \diag \{\A'\W\A\mathbf{1}\}$, $\D_\R  = 2\beta l \lambda_{\max} (\omg' \omg) \I.$									
		\State \textbf{Output:}  $\tilde{\x}^{(I)}$ - reconstructed image.
		\For {i $=0,1,2,\cdots,{I-1}$}		
				
		\State \textbf{(1) Image Reconstruction}: with $ \{\tilde{\z}_j^{(i)}\}$ fixed,	
		
		\textbf{Initialization:}  $\g^{(0)} = M\A_M'\W_M(\A_M\tilde{\x}^{(i)}-\y_M)$, $\ze^{(0)} = \g^{(0)} $, $\h^{(0)} = \D_\A \tilde{\x}^{(i)} - \ze^{(0)}$, $\x^{(0)} = \tilde{\x}^{(i)}$, $\nabla \R (\x^{(0)})= 2 \beta \sum_{j=1}^{N} \big\{ \P_j' \omg' \omg \P_j \tilde{\x}^{(i)} - \P_j' \omg'\tilde{\z}_{j}^{(i)} \big\}$.	
		
		\For {k $=0,1,2,\cdots,K-1$}	
		\For {m $=0,1,2,\cdots,M-1$}	  			
		\begin{equation*}
		\left\{			
		\begin{aligned}
		\s^{(k+1)} &= \rho(\D_\A \x^{(k)} -\h^{(k)}) + (1-\rho)\g^{(k)} \\
		\x^{(k+1)} &= [\x^{(k)} - (\rho\D_\A+\D_\R)^{-1}(\s^{(k+1)} +\nabla \R(\x^{(k)}))]_\mathcal{C} \\
		\ze^{(k+1)} & \triangleq  M \A_m'\W_m(\A_m\x^{(k+1)}-\y_m)  \\
		\g^{(k+1)} &= \frac{\rho}{\rho+1}(\alpha \ze^{(k+1)} + (1-\alpha)\g^{(k)}) +  \frac{1}{\rho+1}\g^{(k)}\\
		\h^{(k+1)}  &= \alpha(\D_{\A} \x^{(k+1)} -\ze^{(k+1)}) + (1-\alpha)\h^{(k)} 
		\end{aligned}
		\right.
		\end{equation*}  
		 decrease $\rho$ using \eqref{eq:rho} with $n = kM +m$.
		\EndFor	
		\EndFor		
		\State   $\tilde{\x}^{(i+1)} = \x^{(K)}$ 
		\State \textbf{(2) Sparse Coding}: with $\tilde{\x}^{(i+1)}$ fixed, the optimal sparse codes are
		$\tilde{\z}_j^{(i+1)} =  H_{\gamma}(\omg \P_j \tilde{\x}^{(i+1)}) \,  \forall \, j$. 
		\EndFor				
	\end{algorithmic}
\end{algorithm}

	Algorithm \ref{alg: PWLS-ST Algorithm} describes the proposed optimization algorithm for solving the PWLS-ST reconstruction problem \eqref{eq:P0}.

\section{Experimental Results}
\label{sec:result}

We evaluate the proposed PWLS-ST method and compare its image reconstruction quality with those of conventional FBP with a Hanning window,  PWLS reconstruction with regularization based on DCT in \eqref{eq:Rx} (PWLS-DCT), and PWLS reconstruction with edge-preserving hyperbola regularization (PWLS-EP). The PWLS-EP reconstruction is optimized using relaxed OS-LALM algorithm \cite{nien:16:rla}. 

 \vspace{-0.15cm}

\begin{figure}[h]
	\begin{minipage}[]{0.49\linewidth}
		\centering
		\centerline{ \subfigure[DCT]{
				\includegraphics[width=2.6cm]{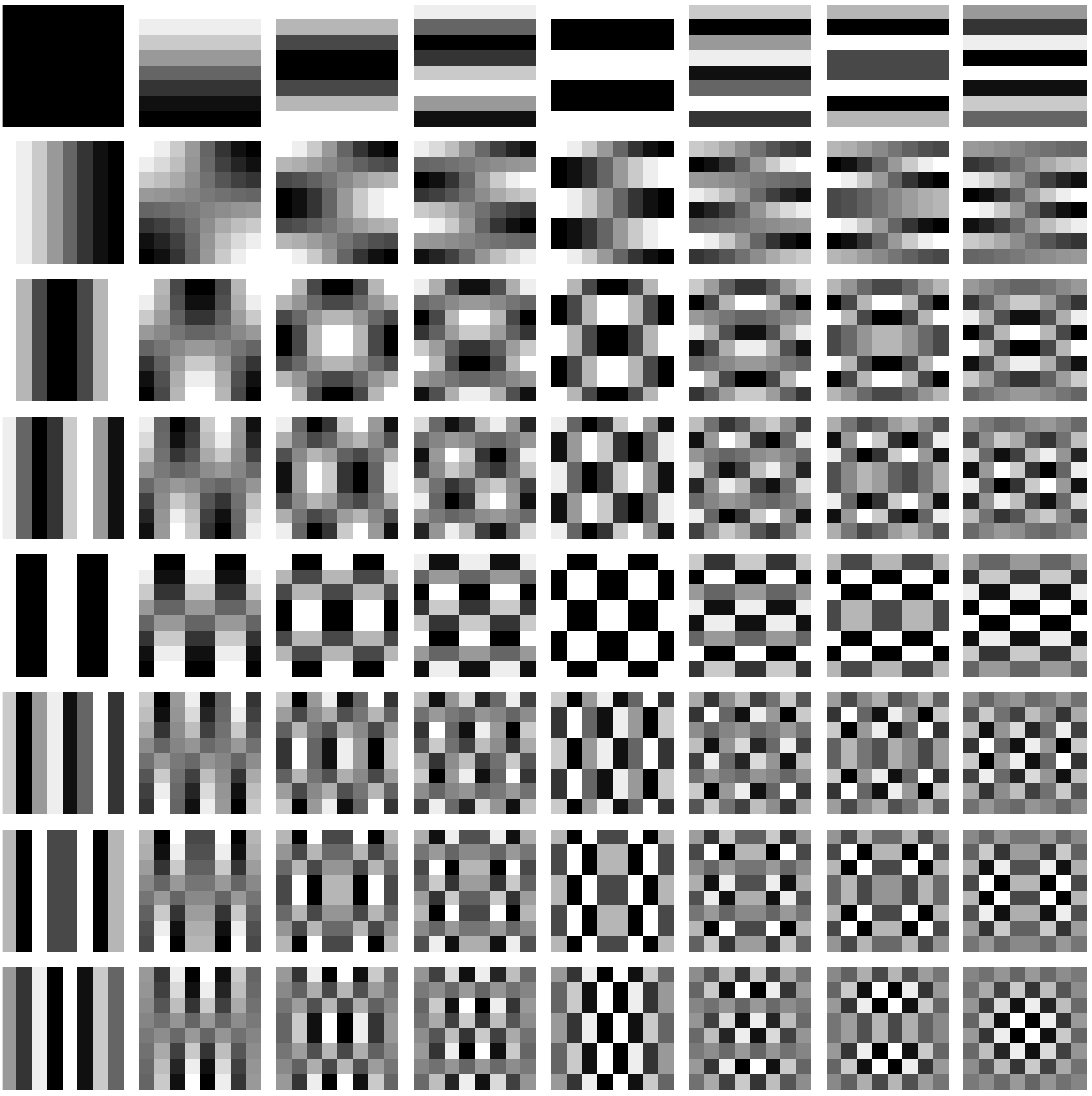}
			}}
		\end{minipage}	
		\begin{minipage}[]{0.49\linewidth}		
			\centerline{ \subfigure[$\eta = 75$]{
					\includegraphics[width=2.6cm]{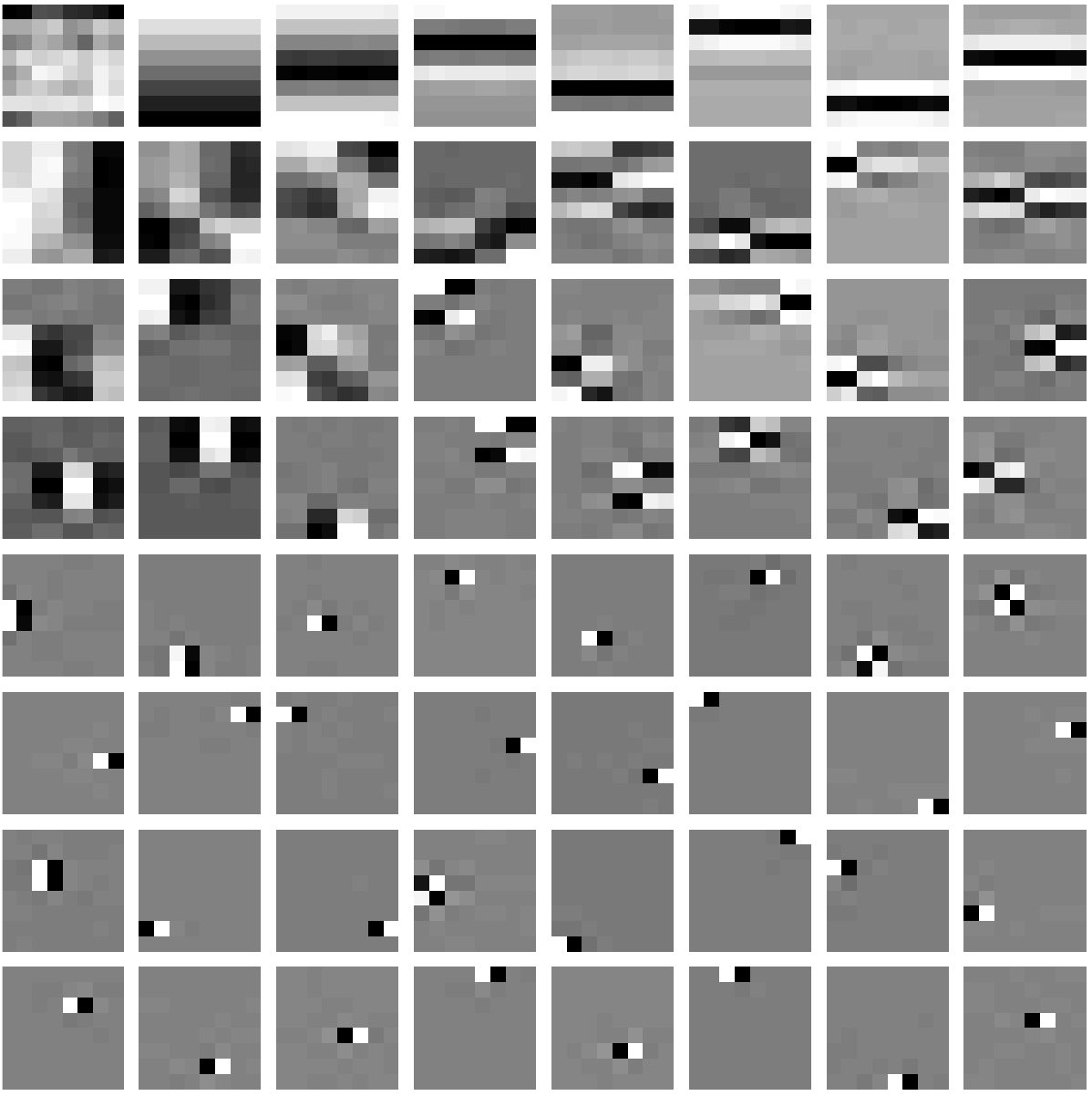}
				}}
			\end{minipage}	
			 \vspace{-0.3cm}
			\caption{Rows of the sparsifying transform shown as $8\times 8$ patches for (a) 2D DCT, and (b) learned $\omg$ with $\eta = 75$.} 
			\label{fig:lear_tran}
			
		\end{figure}

 \vspace{-0.1cm}

We pre-learned a ST matrix from $5$ different slices of an XCAT phantom \cite{segars:08:rcs} using \eqref{eq:P1}. We extracted $8 \times 8$ image patches with overlapping stride of 1 from the five $512 \times 512$ XCAT slices. We ran $2000$ iterations of the alternating minimization algorithm proposed in \cite{ravishankar:15:lst} to make sure the learned ST is completely converged with $\lambda= 5.85\times10^{15}$, and $\eta = 75$. \mbox{Figure \ref{fig:lear_tran}} shows 2D DCT and the well-conditioned pre-learned ST. Each row of these transforms is displayed as an $8 \times 8$ patch.

We simulated a 2D fan-beam CT scan using a $1024 \times 1024 $ XCAT phantom slice, which is different from the learning slices, and $\Delta_x=\Delta_y=0.4883$ mm. Noisy (Poisson noise) sinograms of size $888 \times 984 $ were numerically generated with GE LightSpeed fan-beam geometry corresponding to a monoenergetic source with $10^5$, $10^4$, $5\times 10^3$ and $10^3$ incident photons per ray and no scatter, respectively. We reconstructed a $512 \times 512$ image with a coarser grid, where $\Delta_x=\Delta_y=0.9766$ mm.

\begin{figure*}[t]
     	\centering
     	\begin{tabular}{cc}
     		\includegraphics[width=0.64\linewidth]{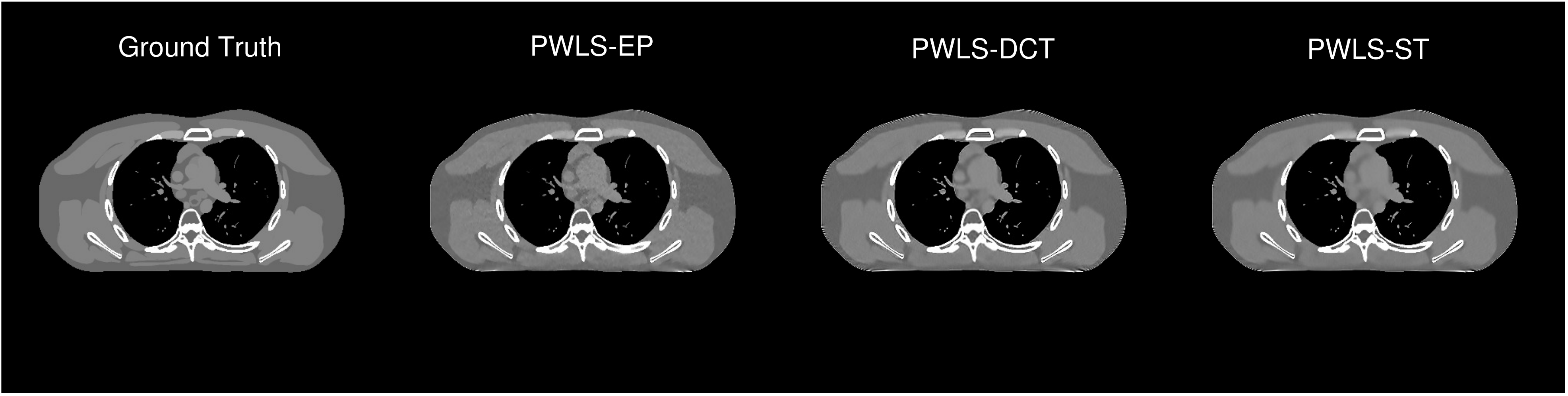} &\includegraphics[width=0.335\linewidth]{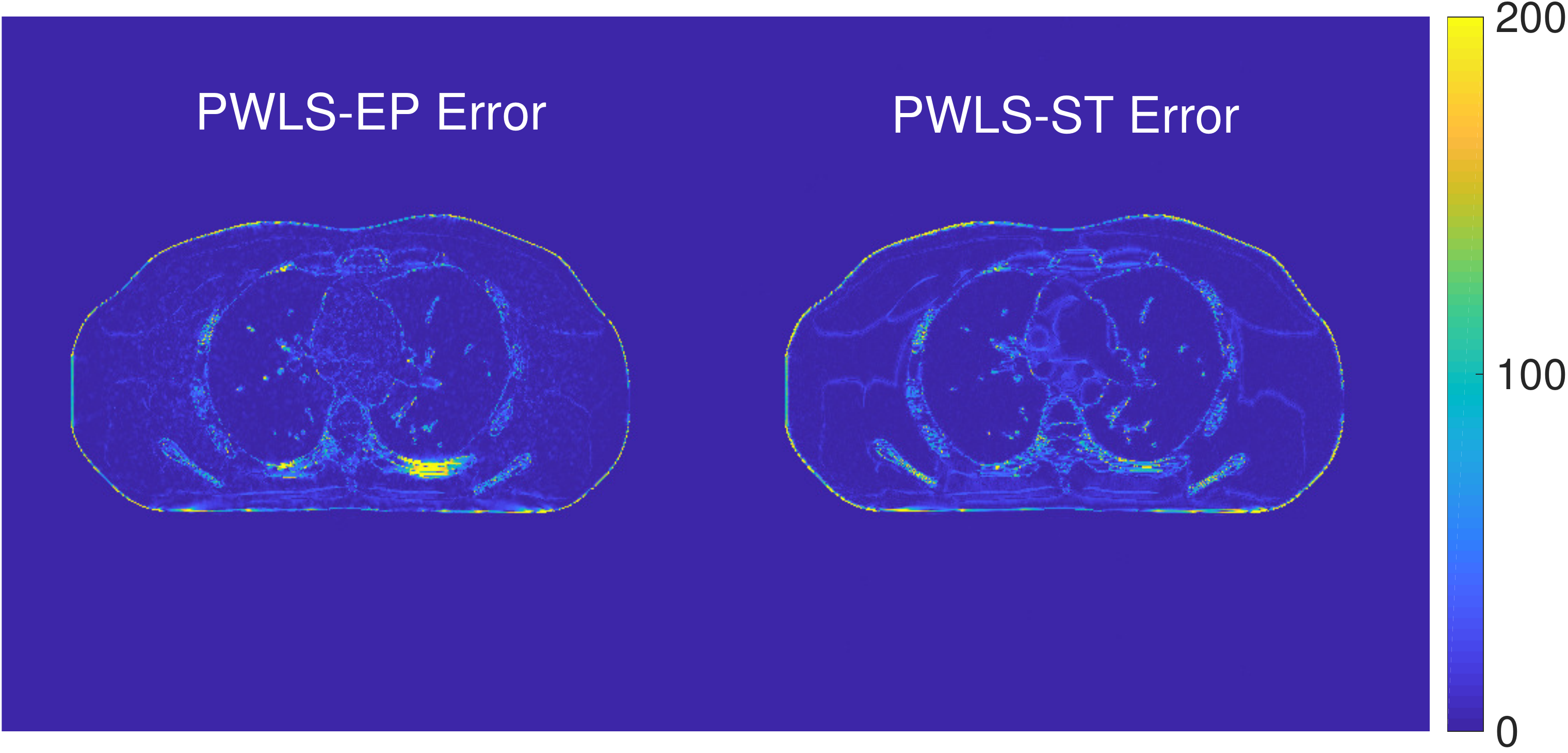} \\		
	 	   \includegraphics[width=0.64\linewidth]{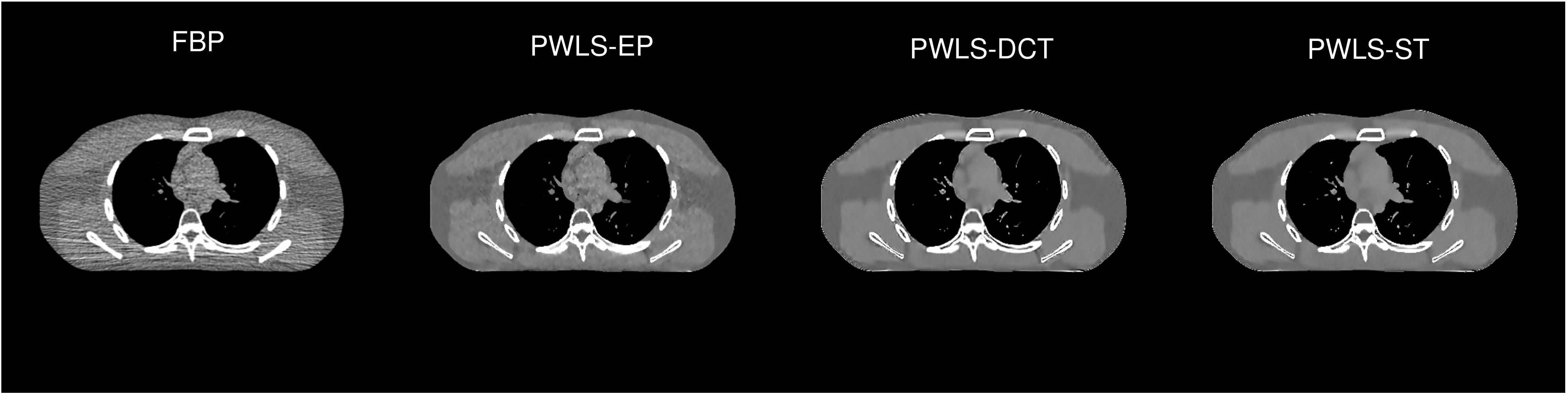} &\includegraphics[width=0.335\linewidth]{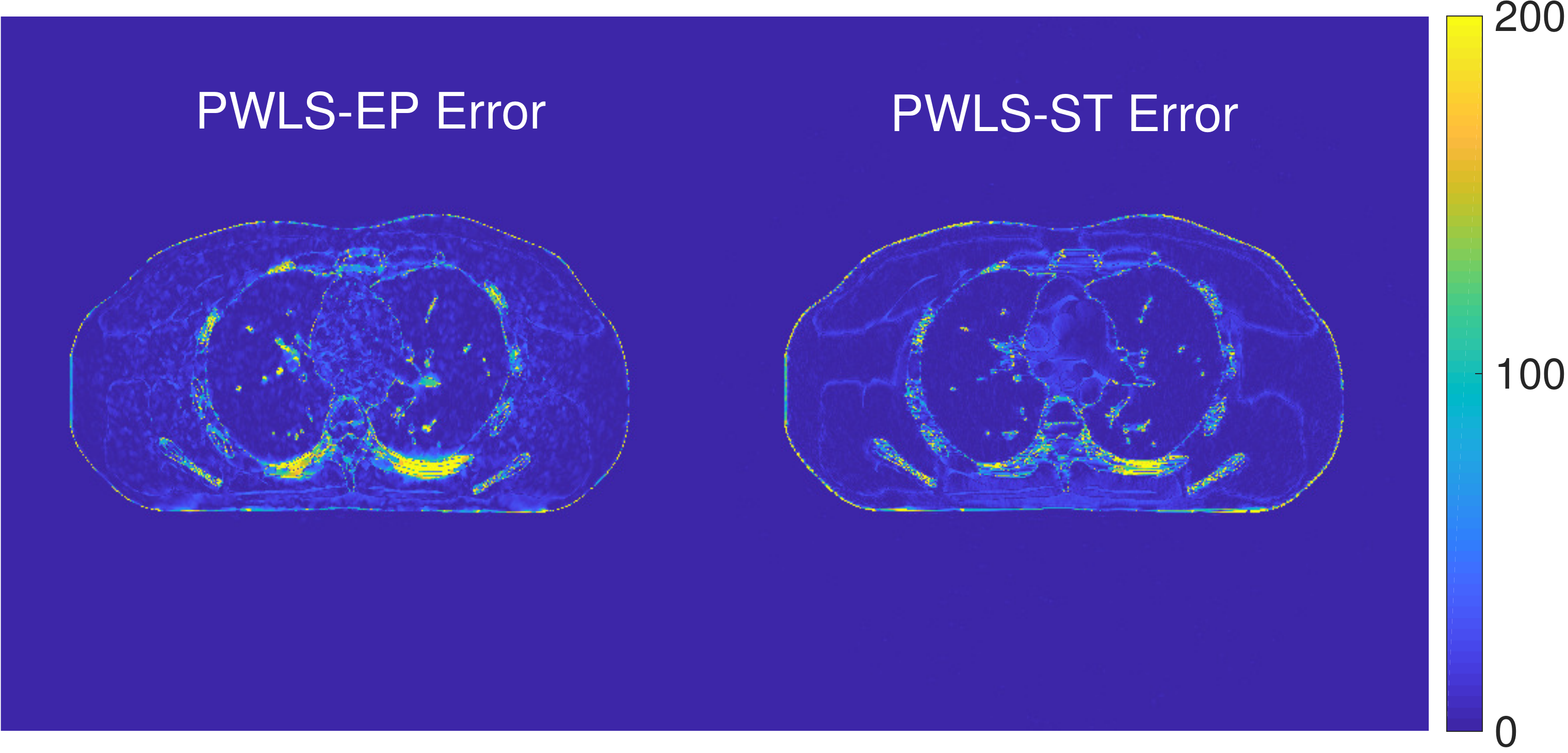} \\
	 	   \includegraphics[width=0.64\linewidth]{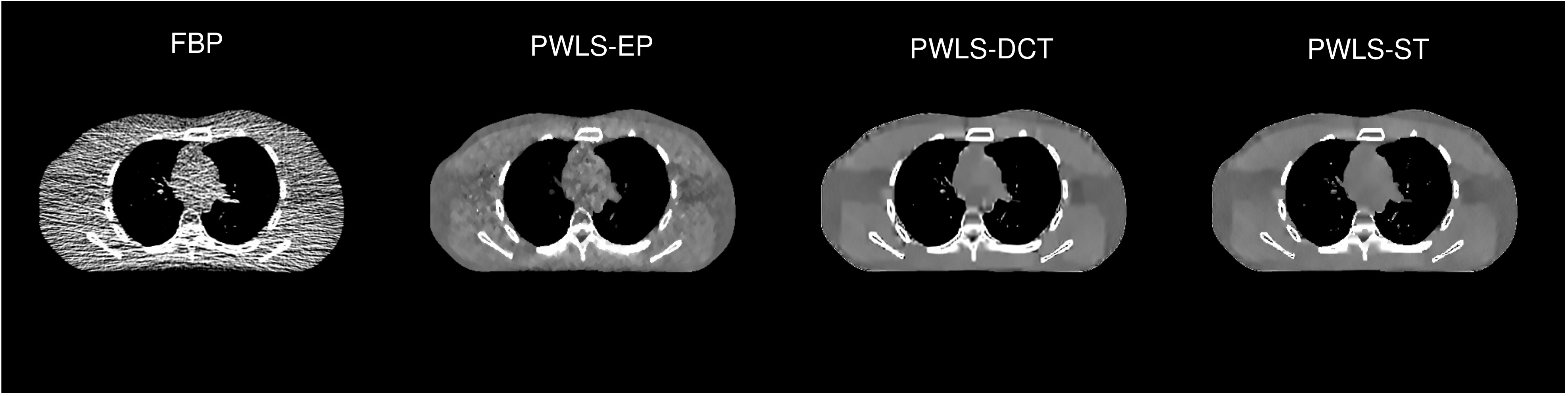} &\includegraphics[width=0.335\linewidth]{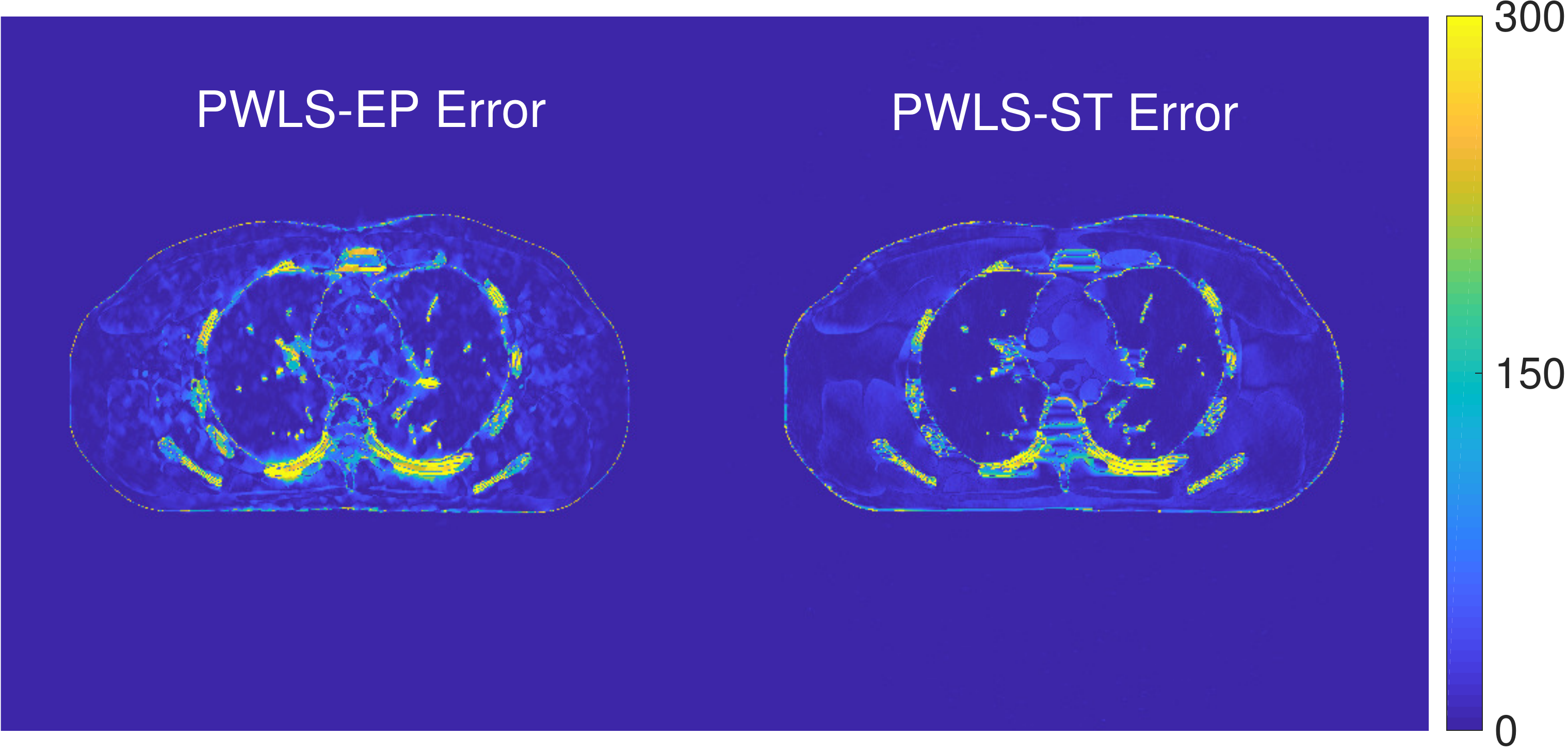} \\
   	    {(a)}  &   {(b) }
        \end{tabular}
           \vspace{-0.3cm}
      	\caption{From top to bottom the incident photon intensities are $10^5$, $10^4$, and $10^3$. (a) Reconstructions (display window $[800, 1200]$ HU). (b) Difference images (magnitudes) between PWLS-EP and PWLS-ST reconstructions and the ground truth.} 
      	\label{fig:image_comp}
\end{figure*}

We computed the root mean square error (RMSE) in modified Hounsfield units (HU) where air is $0$ HU. For a reconstructed image $\hat{\x} \in \mathbb{R}^{N_p}$, RMSE $\triangleq \sqrt{\sum_{i=1}^{N_p}(\hat{x}_i-x^*_i)^2/{N_p}}$, where $\x^*$ is the ground truth image. 

Initialized with FBP reconstructed images, the PWLS-EP method converges quickly using relaxed OS-LALM with $12$ subsets. For PWLS-DCT and the proposed PWLS-ST methods, we used the image obtained with the PWLS-EP method as initialization.
The parameter $\beta$ and $\gamma = 25$ were determined empirically by sweeping over a large range of values. For PWLS-ST, $\beta$ was set as $5.0\times10^{5}$, $1.5\times10^{5}$, $7.0\times 10^{4}$ and $6.0\times 10^{4}$ for incident photon intensities of $10^5$, $10^4$, $5\times10^3$ and $10^3$, and the corresponding $\beta$ values for PWLS-DCT were $2.0\times10^{5}$, $9.0\times10^{4}$, $7.0\times 10^{4}$ and $5.0\times 10^{4}$, respectively. 
In each outer iteration, we ran $2$ inner iterations of the image update step with $4$ subsets, i.e.,  $K = 2$, $M=4$ in Algorithm \ref{alg: PWLS-ST Algorithm}. For PWLS-EP, $\beta$ was set as $2^{11}$, $2^{13}$, $2^{13.5}$ and $2^{15}$ respectively for the four incident photon intensities, and the edge-preserving regularizer is ${\varphi}{(t)}\triangleq\delta^2\left(  \sqrt{1+| t/\delta |^2 }  -1  \right)$ with $\delta=10$ HU.

\begin{table}[h]	
	\centering
	\caption{RMSE (HU) of reconstructions with FBP, PWLS-EP, PWLS-DCT and PWLS-ST for four levels of incident photon intensities. }
	\label{tab:}	
			
	\begin{tabular}{|c|c|c|c|c|}
		\hline
		Intensity                 &    FBP             &  PWLS-EP                    &  PWLS-DCT          & PWLS-ST  \\ \hline
		$1\times10^5$        &  $41.3$         & $\mathbf{19.1}$         &  $ 21.8$              &   $19.2$  \\ \hline  
		$1\times10^4$        &  $47.3$         & $27.9$                        &   $ 27.7 $             &   $\textbf{24.8}$ \\ \hline
		$5\times10^3$        &  $55.3 $       & $32.3$                        &   $31.8$               &  $\textbf{30.6} $ \\ \hline	
  	$1\times10^3$        &  $101.9 $      & $46.7$                        &   $44.4$               &  $\textbf{44.2} $ \\ \hline	
	\end{tabular}

\end{table}

Figure \ref{fig:image_comp}(a) shows the reconstructed images by FBP, PWLS-EP, PWLS-DCT and PWLS-ST. When the incident photon intensity is $10^5$, although the PWLS-EP image has lower RMSE than the PWLS-ST image, the latter has no visible noise. When the incident photon intensities are $10^4$ and $10^3$, compared with FBP and PWLS-EP, PWLS-ST greatly improves image quality in terms of decreasing noise and retaining small structures. Figure \ref{fig:image_comp}(b) shows the difference images (magnitudes) between the PWLS-EP and PWLS-ST reconstructions and the ground truth image, for the three incident photon intensities.

Table \ref{tab:} summarizes the RMSE of reconstructions with FBP, PWLS-EP, PWLS-DCT and PWLS-ST for four different incident photon intensities. For low dose cases, PWLS-ST further decreases the RMSE achieved by PWLS-EP.  


\section{Conclusion}

 We present a PWLS-ST method that combines conventional PWLS reconstruction with regularization based on a sparsifying transform that is pre-learned from a dataset of numerous CT images, to improve the quality of reconstructed images in low dose CT imaging. Numerical experiments show the proposed \mbox{PWLS-ST} method may help reduce X-ray dose to a low level while still providing high quality image reconstructions. For future work, we will investigate PWLS with a union of  sparsifying transforms \cite{wen:14:sos} that could be pre-learned in an online manner \cite{ravishankar:15:ost} from large datasets. We also plan to compare our methods to the recent transform blind reconstruction framework \cite{pfister:14:ast}. We will apply the proposed PWLS-ST method to clinical CT data.

\pagebreak

\bibliographystyle{IEEEbib}
\bibliography{refs}

\end{document}